\documentclass{svproc}
\usepackage{url}

\usepackage[dvips]{graphicx}
\usepackage[latin1]{inputenc}
\usepackage{amssymb,amsmath,array}
\usepackage[dvipsnames]{xcolor} 
\usepackage{subcaption}
\usepackage{float}
\usepackage{multirow}
\usepackage{comment}
\usepackage{siunitx}
\usepackage{tablefootnote}
\usepackage{gensymb}
\usepackage[dvipsnames]{xcolor}
\usepackage{amsmath}
\usepackage{amsfonts} 
\usepackage{hyperref}
\usepackage{tikz}
\newcommand*\circled[1]{\tikz[baseline=(char.base)]{
            \node[shape=circle,draw,inner sep=1pt] (char) {#1};}}
\usepackage[ruled,vlined]{algorithm2e}

\usepackage{cite}  
\usepackage{arydshln} 
\usepackage{dblfloatfix} 

\begin{document}
\mainmatter              
\title{From Elevation Maps To Contour Lines: \\ SVM and Decision Trees to Detect \\ Violin Width Reduction}
\titlerunning{From Elevation Maps To Contour Lines}  
%
\author{Phil\'emon Beghin\inst{1} \and Anne-Emmanuelle Ceulemans\inst{2,3} \and Fran\c{c}ois Glineur\inst{1}}
\authorrunning{Phil\'emon Beghin et al.} 
%
%

\institute{UCLouvain - ICTEAM - Louvain-la-Neuve, Belgium
\and
UCLouvain - INCAL - Louvain-la-Neuve, Belgium 
\and
Musical Instruments Museum (MIM) - Brussels, Belgium}

\maketitle              
\setcounter{footnote}{0} 

\begin{abstract}
We explore the automatic detection of violin width reduction using 3D photogrammetric meshes. We compare SVM and Decision Trees applied to a geometry-based ``raw" representation built from elevation maps with a more targeted, feature-engineered approach relying on parametric contour lines fitting. Although elevation maps occasionally achieve strong results, their performance does not surpass that of the contour-based inputs. 
\end{abstract}

\section{Historical Context of Violin Reduction and Motivation}

\begin{figure}
  \centering
  \begin{subfigure}{0.24\linewidth}
    \centering
    \includegraphics[scale = 0.115]{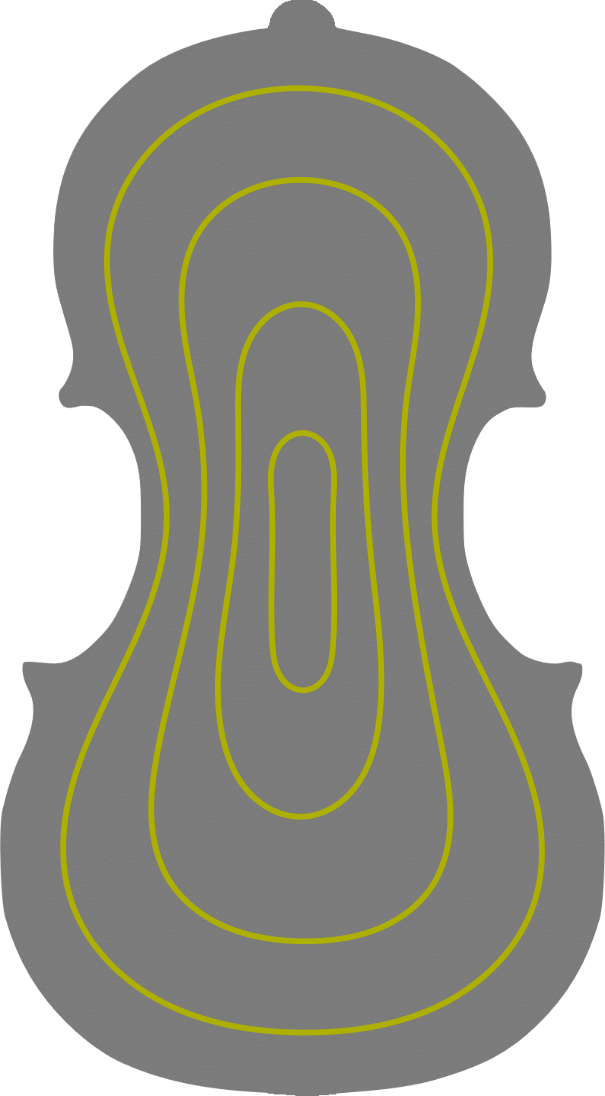}
  \end{subfigure}%
  \begin{subfigure}{0.24\linewidth}
    \centering
    \includegraphics[scale = 0.115]{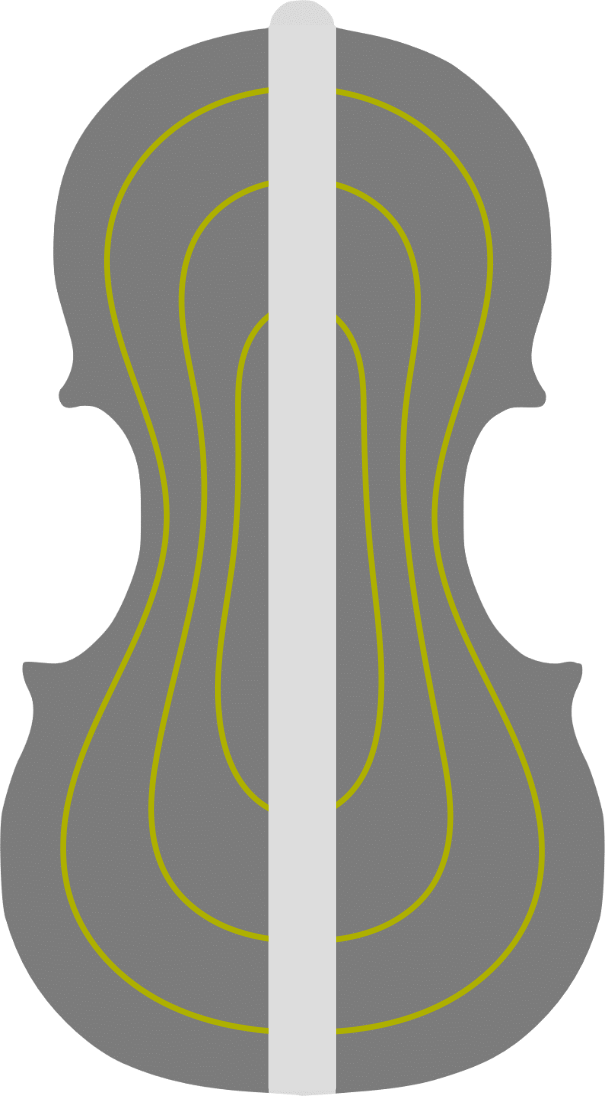}
  \end{subfigure}%
  \begin{subfigure}{0.24\linewidth}
    \centering
    \includegraphics[scale = 0.18]{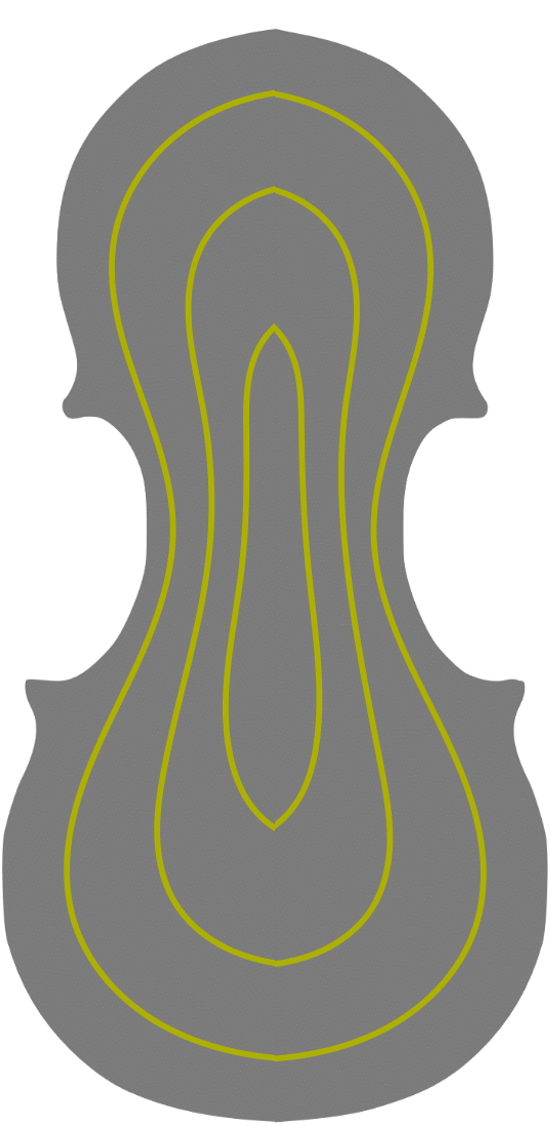}
  \end{subfigure}
  \caption{Impact of the reduction of the width of the sound box on the contour lines.}
  \label{fig:Reduction CL}
\end{figure}

Some historical violins built before 1750 have been reduced at some point of their history to match standard sizes \cite{ceulemans2023baroque}. A possible kind of reduction consisted in removing a slice of wood along the central axis of the instrument, and then gluing the two halves back together, to diminish its width, as observed in \cite{radepont2020revealing}. This has a direct impact on the contour lines, which are rather ``U-shaped" before reduction, as in Figure \ref{fig:Reduction CL} (left) and become ``V-shaped" afterwards (right). In museum and restoration practice, luthiers traditionally rely on visual inspection, tactile assessment, and endoscopic examination to identify possible reductions. These approaches, while invaluable, remain subjective and may overlook tiny geometric evidence, particularly in instruments that have undergone partial deformation due to ageing, cracks, or natural warping of wood.

\setlength{\belowcaptionskip}{-10pt}

\noindent Therefore quantitative analyses that can objectively characterise the geometry of historical instruments are strongly needed. Determining whether a historical violin has been reduced is essential for understanding its true geometry, authenticity, and cultural context. Such quantitative analysis supports more reliable organological interpretations, conservation decisions, and historical reconstructions than traditional expert-based assessment. Our goal is to develop an objective and systematic approach capable of assisting with this classification in a reproducible manner.

\section{Data Acquisition}

\noindent Our corpus consists in 25 violins and violas from the Musical Instruments Museum (Brussels), and it is strongly suspected that $5$ of them are reduced. All sound board geometries were obtained through a validated photogrammetric workflow achieving sub-millimetre accuracy. The acquisition pipeline and quantitative validation are detailed in \cite{beghin2023validation}. 

\subsection{Elevation Maps}
\label{section:elevation_maps}
We first represent the geometry of each instrument's sound board using two-dimensional elevation maps. We derive them by computing the heights of a 3D mesh surface over a square grid of regularly spaced points (every \SI{0.25}{\mm}). This grid is parallel to the plane of symmetry of the violin, as defined in \cite{beghin2026discussion}. Nodes of the grid which do not intersect the mesh are filled with zeros (instead of e.g. \texttt{NaN} values), which corresponds to the elevation of the plane of symmetry. Even though it is usually not recommended to fill missing data with zero values, this makes sense in this particular application since it is aligned with the horizontal plane with zero elevation, as described in \cite{beghin2026discussion}.\\

\begin{figure}
    \centering
    \includegraphics[scale=0.5]{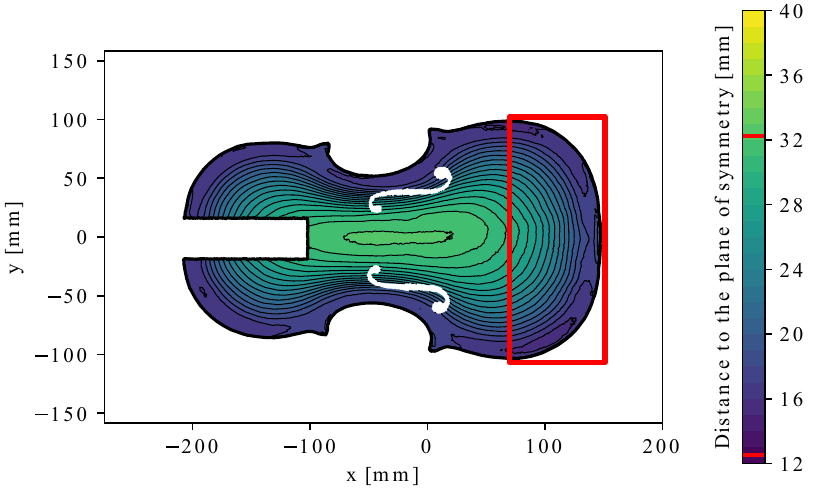}
    \caption{Elevation map sampled every \SI{0.25}{\mm} $\times$ \SI{0.25}{\mm} and zone of interest (red rectangle). The two red lines on the scale bar indicate the minimum and maximum elevations of the violin sound board.}
    \label{fig:elevation map}
\end{figure}

\begin{figure}
\centering
\begin{subfigure}[b]{0.475\textwidth}
\hspace*{-0.7cm}\includegraphics[height=4.2cm]{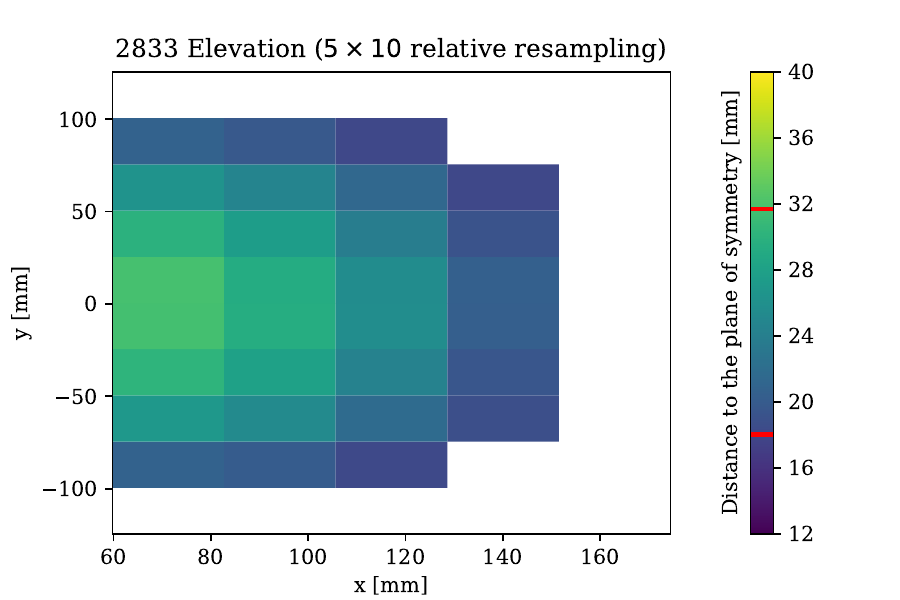}
\vspace{\baselineskip}
\hspace*{-0.7cm}\includegraphics[height=4.2cm]{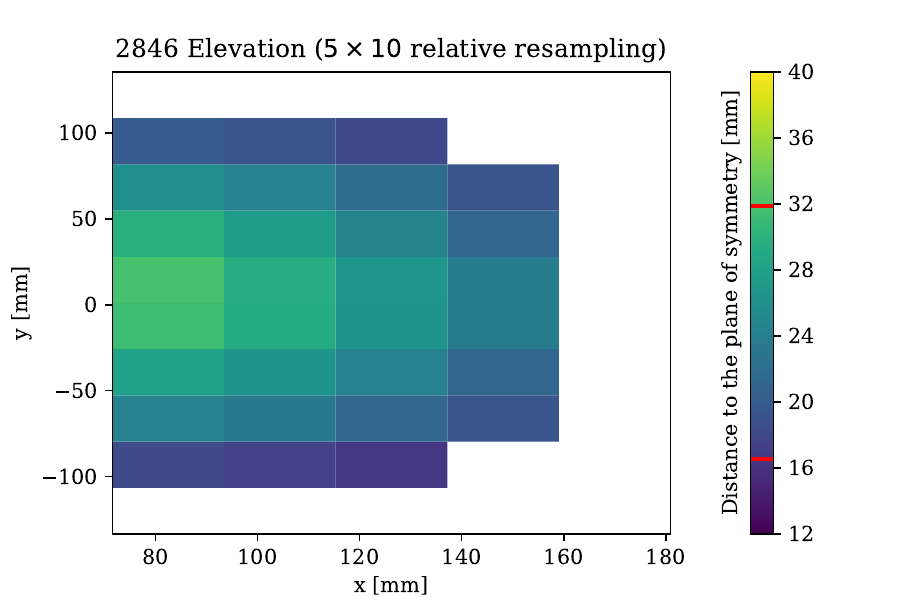}
\end{subfigure}%
\begin{subfigure}[b]{0.475\textwidth}
\includegraphics[height=4.2cm]{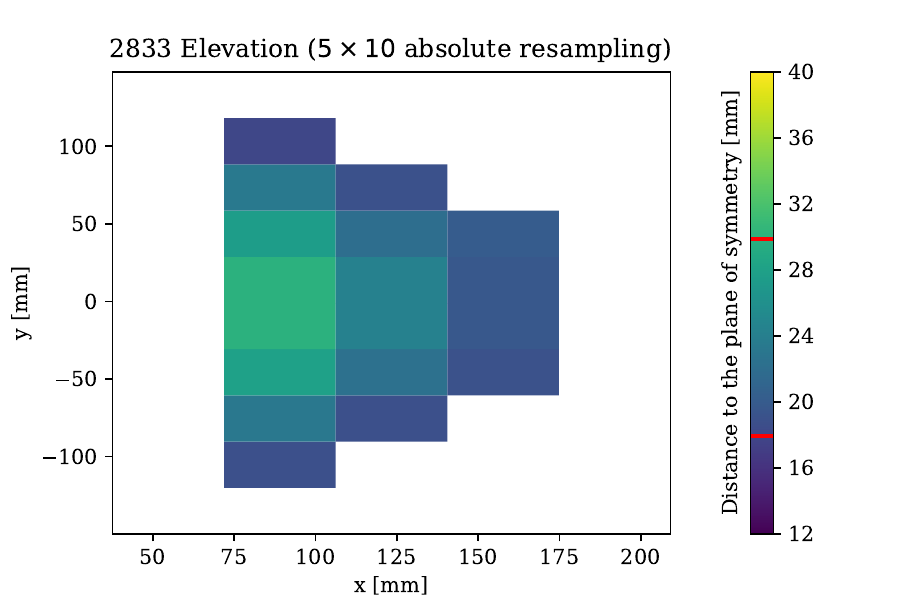}
\vspace{\baselineskip}
\includegraphics[height=4.2cm]{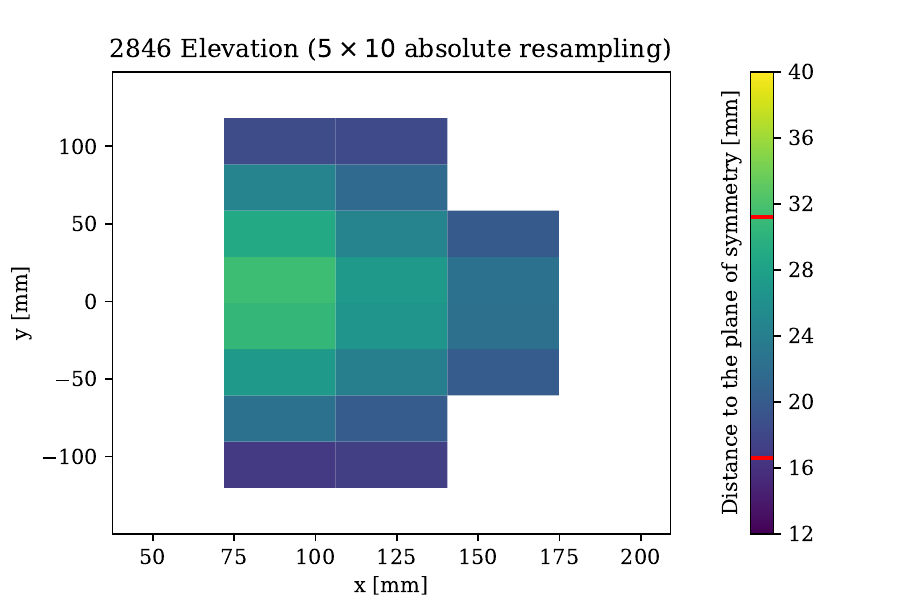}
\end{subfigure}%
\caption{Elevation maps resampled on a $5 \times 10$ square grid. The top line shows the resampling of a non reduced instrument (inv. no 2833) while the bottom line illustrates the resampling of a reduced instrument (inv. no 2846). The left-hand side highlights relative resampling and the right-hand side absolute resampling. Zones with zero elevation (no intersection between the grid and the mesh) are left in white. The two red lines on each scale bar indicate the minimum and maximum elevations of the violin sound boards.}
\label{fig:5_10_grids}
\end{figure}

\noindent In order to focus on similar zones as in the parameter profile-based approach below, we restrict the elevation maps to the bottom of the sound board; precisely starting from where it is the widest, as illustrated in red in Figure \ref{fig:elevation map}. Since SVM input vectors must share the same dimension, we resample the elevation maps in two different ways. In the \emph{relative} approach, each instrument's bounding box is subdivided proportionally, so that all resulting grids have identical dimensions. The bounding box dimensions across the whole violin dataset are approximately \SI{100}{\mm} $\times$ \SI{250}{\mm}. When considering all instruments, the length-to-width ratio varies between 2.266 (smallest ratio) and 2.765 (largest ratio). To ensure a consistent and representative spatial discretisation, we therefore adopt a resampling grid with a ratio close to 2.5, as reported in Tables \ref{tab:SVM} and \ref{tab:Decision Trees}. The coarsest resolution ($5 \times 10$ cells) corresponds to a grid of squares of approximately \SI{20}{\mm} $\times$ \SI{20}{\mm}, and the resampling is progressively refined up to a \SI{1}{\mm} $\times$ \SI{1}{\mm} square grid ($100 \times 250$ cells). In the second, \emph{absolute} approach, we first define a common ``large" bounding box for all instruments, discretised at a fixed spatial resolution of 0.25 mm; each elevation map is then extracted within this global frame. These absolute maps are subsequently downsampled to various grid resolutions. The length-to-width ratio of the bounding box in the absolute resampling approach is approximately 1.94. However, for consistency across experimental settings, we retained grids with a ratio close to 2.5 in both cases. Relative and absolute resampled grids are shown for reduced and non reduced instruments in Figure \ref{fig:5_10_grids}.

\noindent We test both raw elevations and normalised elevations. For the latter, we perform centering by subtracting their mean and scaling by their standard deviation. Both gobal statistics were computed while excluding undefined entries of the elevation maps which had been set to zero to avoid biasing the corresponding estimates.

\subsection{Parameter Profile-based Approach}

\begin{figure}
    \centering
    \includegraphics[scale=0.4]{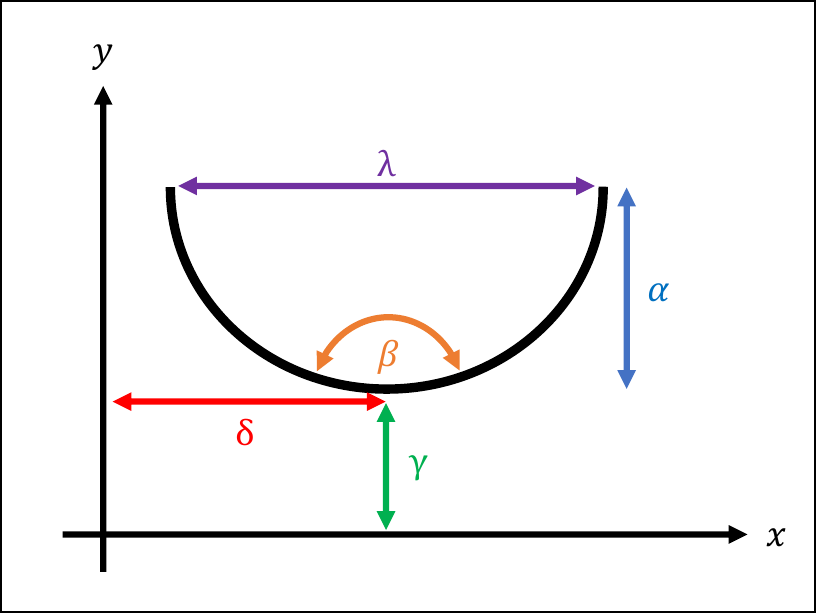}
    \caption{Impact of the parameters on the contour lines fitting.}
    \label{fig:profiles}
\end{figure}

\noindent In \cite{beghin2025identification} we introduced problem-specific geometric features based on contour lines of the violin sound board. From the 3D meshes properly aligned with respect to the plane of symmetry \cite{beghin2026discussion}, we derived a series of contour lines corresponding to successive height levels separated by one millimetre. Each contour line was then fitted with a ``parabola-like" curve described by four parameters $\alpha, \beta, \gamma, \delta$ in $y = \alpha |\frac{x-\delta}{\lambda/2}|^\beta+\gamma$, capturing respectively the vertical stretch/compression, curvature opening, vertical offset, and horizontal offset, as illustrated in Figure \ref{fig:profiles} (width $\lambda$ was measured and therefore not optimised to fit the data, it was used instead to normalise the $x$ dimension \cite{beghin2025identification}). \\

\begin{figure}
    \centering
    \includegraphics[scale=0.4]{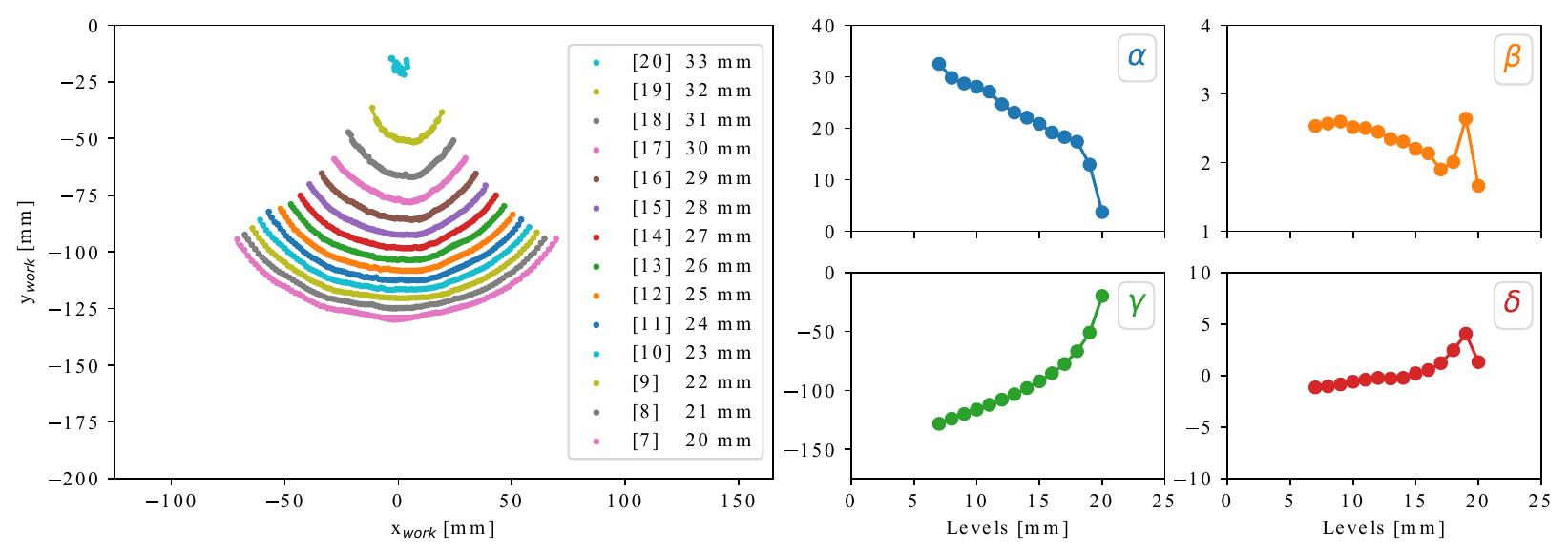}
    \caption{Fitting of contour lines of the sound board of an instrument (left) and profile of the parameters $\alpha$, $\beta$, $\gamma$ and $\delta$ for each level (right).}
    \label{fig:CL and profiles}
\end{figure}

\noindent This produces four parameter profiles ($\pmb{\alpha}$, $\pmb{\beta}$, $\pmb{\gamma}$, $\pmb{\delta}$) across levels, representing a compact numerical signature of each instrument. Figure \ref{fig:CL and profiles} shows the fitted contour lines of a violin sound board on the left, and the parameter profile on the right. Although we will show below that the opening parameter $\beta$ is a very discriminative feature, note that these one-dimensional parameter profiles are very dependent on the quality of the fitted contour lines. Indeed, since our approach was based on the profiles, but also on the general aspect of those (we will approximate them with linear, piecewise-linear and quadratic polynomials), a single outlier can skew values (see  e.g. the spike at the penultimate $\beta$ level in Figure \ref{fig:CL and profiles}).

\subsection{PCA Reduction of Elevation Maps}

While elevation maps provide a dense description of the sound board geometry, their dimensionality rapidly becomes very high when fine spatial resolutions are used (up to $100 \times 250$ cells = 25 000 features). Given the limited size of our corpus, such high-dimensional representations may lead to overfitting and reduce the stability of statistical classifiers. At the same time, the parameter profile-based approach demonstrates that compact, low-dimensional geometric descriptors can already yield strong discriminative performance. This motivates the exploration of an intermediate strategy: reducing the dimensionality of elevation maps while preserving as much geometric information as possible. \\

\noindent To this end, we apply Principal Component Analysis (PCA) to the resampled elevation maps. Each grid is first vectorised into a single feature vector, so that each instrument is represented as a point in a high-dimensional Euclidean space. Since PCA cannot handle missing values, particular care was taken in the preprocessing stage. Whenever a grid node corresponded to a missing value (e.g., due to slight differences in geometry coverage between instruments), this node location was systematically removed from all instruments. In other words, we constructed a common binary mask across the dataset: if a given spatial coordinate was undefined for at least one instrument, it was excluded for every instrument\footnote{This motivated us to apply the PCA only to the relative resampled grids. If we had considered absolute resampling, undefined values would have appeared at too many places due to the instruments located at different places within the `large' grid.}. This guarantees that all input vectors share exactly the same dimensions and correspond to the same spatial support before applying PCA, which is then computed across the full dataset. By projecting each elevation map onto the first $k$ principal components, we obtain a reduced representation consisting of only a few coefficients (e.g., $k = 2, 3, 5, 8, 10,$ or $12$, as listed in Tables \ref{tab:svm_rbf_pca}, \ref{tab:svm_linear_pca} and \ref{tab:decision_tree_pca}), which are subsequently used as input features for classification. \\

\begin{figure}
\centering
\begin{subfigure}[b]{0.475\textwidth}
\hspace*{-0.7cm}\includegraphics[height=4.2cm]{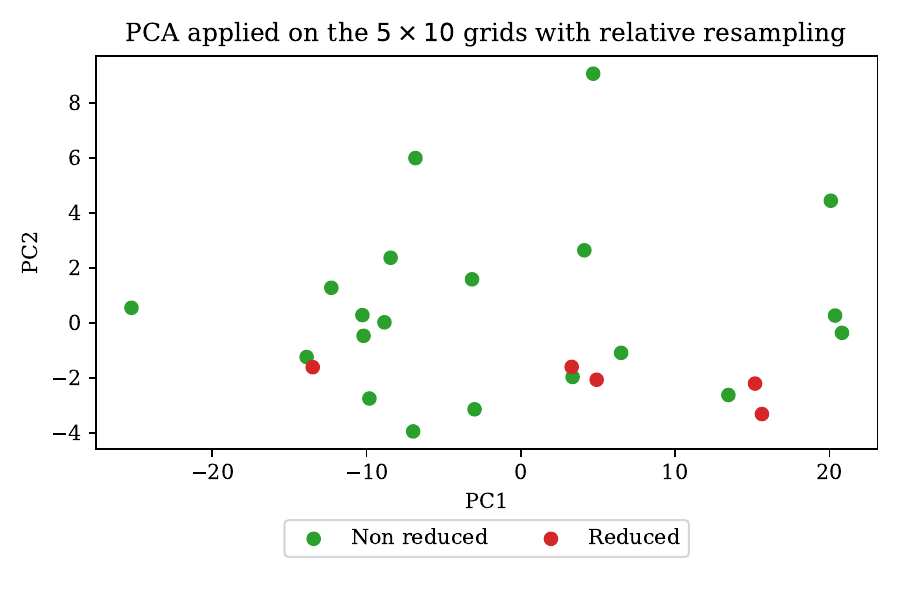}
\vspace{\baselineskip}
\hspace*{-0.7cm}\includegraphics[height=4.2cm]{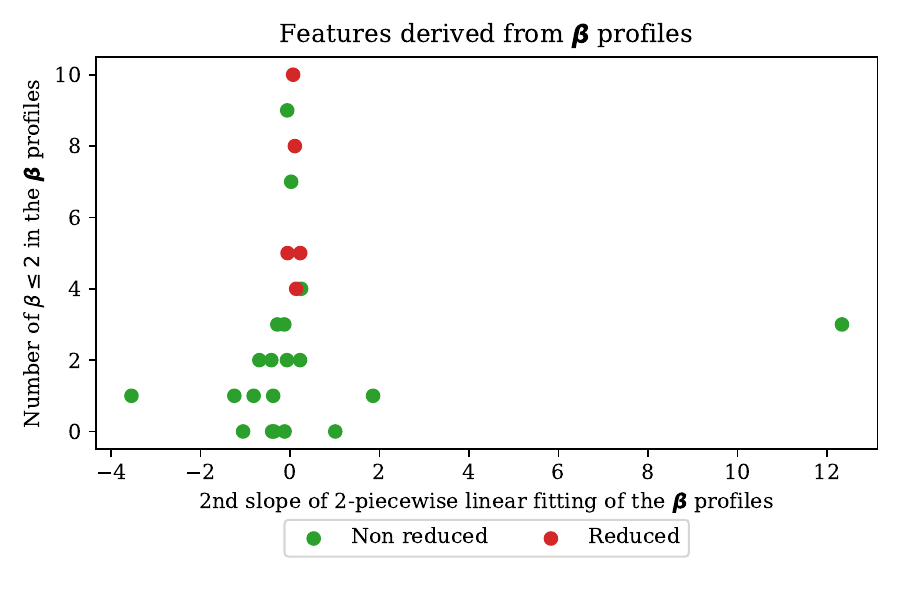}
\end{subfigure}%
\begin{subfigure}[b]{0.475\textwidth}
\includegraphics[height=4.2cm]{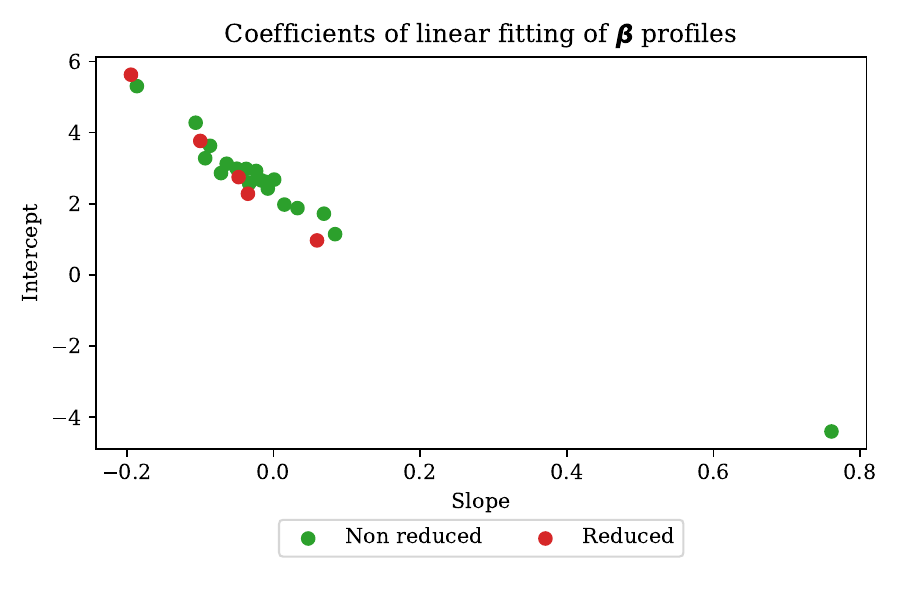}
\vspace{\baselineskip}
\includegraphics[height=4.2cm]{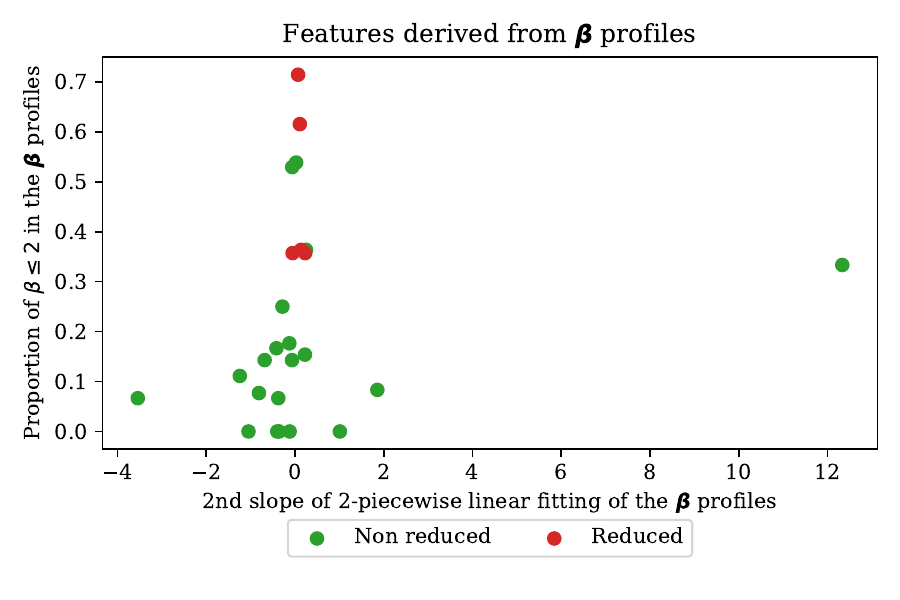}
\end{subfigure}%
\caption{Several 2D representations of violin features. The top-left figure illustrates the first two principal components of a $5\times10$ relative resampled grid (first line of Tables \ref{tab:svm_rbf_pca} and \ref{tab:svm_linear_pca}) and the other three highlight features derived from the $\pmb{\beta}$ profiles (first three lines of the third box of Tables \ref{tab:SVM} and \ref{tab:Decision Trees}).}
\label{fig:2D_visualisation}
\end{figure}

\noindent As opposed to raw elevation maps, which are high dimensional, both the profile-based features and the PCA-reduced maps can produce low-dimensional representations. In Figure \ref{fig:2D_visualisation} we display a few of those, highlighting the two considered classes (reduced vs. non reduced).

\subsection{Other Approaches}
Standard point cloud classification techniques, such as PointNet \cite{qi2017pointnet}, DeepSets\cite{zaheer2017deep}, 
and related architectures were tested in preliminary experiments. They seem to be unable to detect the extremely subtle geometric nuances that differentiate reduced violins from non reduced ones (such as the slight ``U" or ``V" shape of the contour lines). Moreover our dataset is very small (inherent to this research area), and such deep models typically require a much larger corpus. Hence we have to use approaches that can handle a limited number of samples, such as Support Vector Machines (SVM) or Decision Trees. Deep-learning architectures such as multilayer perceptrons or convolutional neural networks, which are an a priori good fit for elevation maps, were not kept due to poor preliminary results, most probably due the small size of the corpus.

\section{Classification}

\subsection{Training Methodology}

As our 25 instruments do not comfortably allow for a typical 70\%-30\% split between training and validation sets, we opt for a leave-one-out cross-validation. We train each of our models on all but one instruments, and then evaluate their performance on the last remaining one. We repeat this 25 times to draw a conclusion about classification accuracy over the whole dataset. Since our two classes have very different sizes, we use balanced accuracy defined as $\frac{TPR + TNR}{2}$, average between the True Positive Rate (TPR) and True Negative Rate (TNR). 

\subsection{Support Vector Machine Classifier}

\begin{figure}
    \centering
    \begin{subfigure}{0.48\textwidth}
    \includegraphics[scale = 0.4]{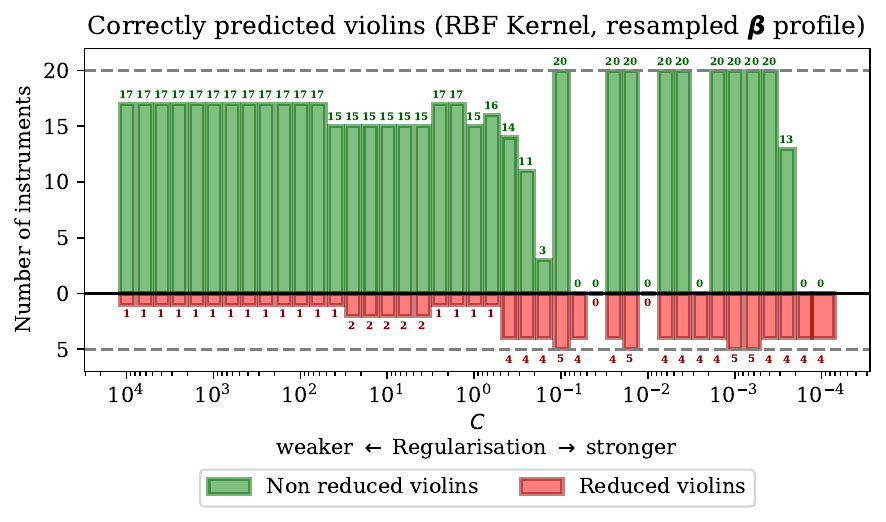}
    \end{subfigure}
    \hfill
    \begin{subfigure}{0.48\textwidth}
    \includegraphics[scale = 0.4]{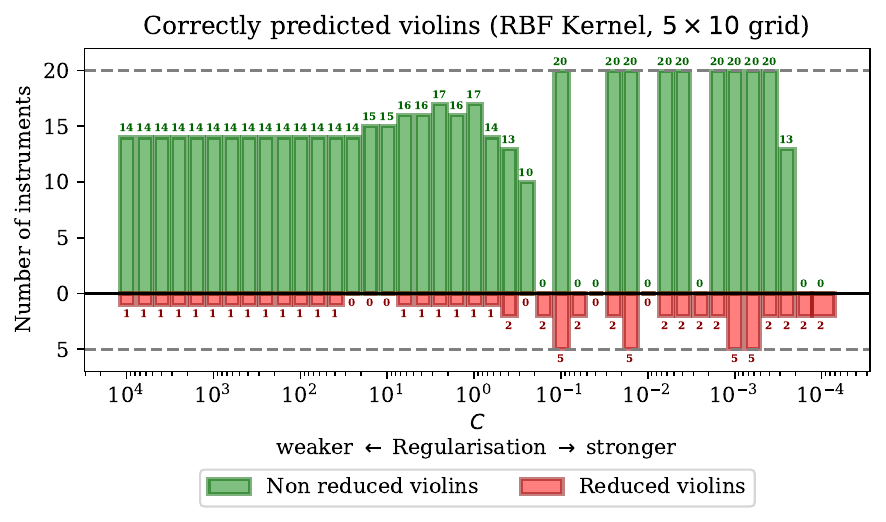}
    \end{subfigure}
    \caption{\small Leave-one-out cross-validation for several regularisation strengths. Input features: resampled $\pmb{\beta}$ profile (left); $5 \times 10$ elevation map with relative resampling (right).}
    \label{fig:C}
\end{figure}

\noindent As in \cite{beghin2025identification} we observe that SVM models are particularly sensitive to the regularisation hyperparameter $C$, which governs the trade-off between maximising the margin and penalising classification errors on the training data. Figure \ref{fig:C}  shows this happens for both the parameter profile-based (left) and the elevation map features (right). Even a slight modification of this parameter can cause drastic fluctuations in performance. Hence, to avoid reporting results for an arbitrary or especially (un)lucky choice of regularisation parameter, we delegate this choice to Algorithm \ref{Algo:23-1-1}, performing an inner cross-validation loop for each leave-one-out split to identify values of $C$ that maximise balanced accuracy. This ensures that our results are meaningful estimates of the generalisation capability.

\begin{algorithm*}[t]
\footnotesize
\SetAlgoSkip{1pt}
\SetAlgoInsideSkip{1pt}
\SetInd{0.5em}{0.5em}
\DontPrintSemicolon

\KwData{25 instruments}
\KwResult{Leave-one-out evaluation and balanced accuracy}

\ForEach{instrument \circled{T} out of the 25}{
    \ForEach{$C \in \{10^{-4}, \dots, 10^4\}$}{
        \ForEach{instrument \circled{V} from the 24 others (excluding \circled{T})}{
            Train the model on the 23 other instruments (excluding \circled{V} and \circled{T})\;
            Validate the model on \circled{V}\;
        }
        Define validation score for this value of $C$ as the balanced accuracy over all \circled{V}s\;
    }
    Select the value of $C$ with the best validation score\;
    Train a new model with this $C$ on the 24 instruments (excluding \circled{T})\;
    Validate the model on instrument \circled{T}\;
}
Compute the overall balanced accuracy on all 25 instruments \circled{T}\;

\caption{\footnotesize Robust regularisation choice with per-instrument cross-validation}
\label{Algo:23-1-1}
\end{algorithm*}

\subsection{Decision Trees}

We additionally investigate Decision Tree classifiers as a complementary approach. Given the small size of the dataset, tree depth must remain strongly constrained to prevent overfitting. As for the regularisation hyperparameter $C$ of the SVM, Algorithm \ref{Algo:23-1-1} carefully selects the optimal tree depth based on balanced accuracy. For the parameter profile-based features and for PCA elevation maps (which both involve a limited number of input parameters), we restrict the maximum depth to values between 1 and 3. For the full elevation map representations, which involve a substantially higher-dimensional feature space, we extend this range up to a maximum depth of 5. 

\section{Classification Results}

\noindent Our aim is to compare performances of SVMs and Decision Trees with the high dimensional elevation maps, their lower dimensional PCA projection and the structured parameter profile-based approach. We want to determine whether a raw input (elevation maps) can perform as well as domain-specific, engineered features (contour line parameters). \\

\subsection{SVM classifiers}

\noindent Recall that we test both \emph{relative} and \emph{absolute} resamplings. All SVM classifiers are computed with a Radial Basis Function (RBF) and with a Linear kernel. In addition, we test the use of either normalised or non-normalised heights in elevation maps (see Section \ref{section:elevation_maps}). Finally, we test different procedures to select the regularisation parameter $C$ in Algorithm \ref{Algo:23-1-1} in cases several best candidate values are tied. We test choosing either the largest or the smallest value of $C$ among the best candidates (corresponding respectively to weakest or strongest regularisation). Since our corpus is small and our classes are unbalanced (20 reduced instruments vs. 5 unreduced ones), we specifically set our SVMs models to take into account the class unbalance through the dedicated hyperparameter, and we assess the quality of the classification using the balanced accuracy metric. \\

%
%

\begin{table*}[h!]
\centering
\resizebox{\textwidth}{!}{
\begin{tabular}{|l|cc|cc||cc|cc|}
\hline
& \multicolumn{8}{c|}{\textbf{Balanced accuracy} (in $\%$)} \\

\textbf{Number and type of features} 
    & \multicolumn{2}{c|}{\textbf{Max $C$}} 
    & \multicolumn{2}{c||}{\textbf{Min $C$}} 
    & \multicolumn{2}{c|}{\textbf{Max $C$}} 
    & \multicolumn{2}{c|}{\textbf{Min $C$}} \\
    & \multicolumn{2}{c|}{\textbf{No norm.}} 
    & \multicolumn{2}{c||}{\textbf{No norm.}} 
    & \multicolumn{2}{c|}{\textbf{Norm.}} 
    & \multicolumn{2}{c|}{\textbf{Norm.}} \\

    & RBF & Lin. & RBF & Lin. & RBF & Lin. & RBF & Lin. \\
\hline
50 ($5 \times 10$ relative grid resampling)       & 20 & 57.5 & 70 & 57.5 & 20 & \textbf{90} & 70 & \textbf{90} \\
250 ($10 \times 25$ relative grid resampling)     & 30 & 60 & 80 & 60 & 20 & 80 & 70 & 80 \\
1000 ($20 \times 50$ relative grid resampling)    & 40 & 45 & \textbf{90} & 42.5 & 20 & 67.5 & 70 & 67.5 \\
1625 ($25 \times 65$ relative grid resampling)    & 40 & 37.5 & \textbf{90} & 35 & 20 & 72.5 & 70 & 72.5 \\
6250 ($50 \times 125$ relative grid resampling)   & 30 & 37.5 & 80 & 37.5 & 20 & 52.5 & 70 & 52.5 \\
14250 ($75 \times 190$ relative grid resampling)  & 40 & 47.5 & \textbf{90} & 47.5 & 20 & 52.5 & 70 & 52.5 \\
25000 ($100 \times 250$ relative grid resampling) & 30 & 40 & 80 & 40 & 20 & 50 & 70 & 52.5 \\
\hline
50 ($5 \times 10$ absolute grid resampling)       & 0  & 47.5 & 50 & 45 & 30 & \textbf{90} & 80 & \textbf{90} \\
250 ($10 \times 25$ absolute grid resampling)     & 0  & 60 & 50 & 50 & 40 & \textbf{90} & \textbf{90} & \textbf{90} \\
1000 ($20 \times 50$ absolute grid resampling)    & 20 & 62.5 & 70 & 62.5 & 30 & 65 & 80 & 65 \\
1625 ($25 \times 65$ absolute grid resampling)    & 20 & 62.5 & 70 & 62.5 & 30 & 70 & 80 & 70 \\
6250 ($50 \times 125$ absolute grid resampling)   & 20 & 47.5 & 70 & 47.5 & 20 & 65 & 70 & 65 \\
14250 ($75 \times 190$ absolute grid resampling)  & 20 & 50 & 70 & 50 & 30 & 70 & 80 & 70 \\
25000 ($100 \times 250$ absolute grid resampling) & 10 & 45 & 60 & 45 & 30 & 50 & 80 & 50 \\
\hline
\end{tabular}
}

\vspace*{.2cm}

\resizebox{\textwidth}{!}{
\begin{tabular}{|l|cc|cc|}
\hline
& \multicolumn{4}{c|}{\textbf{Balanced accuracy} (in $\%$)} \\

\textbf{Number and type of features} 
    & \multicolumn{2}{c|}{\textbf{Max $C$ (weak reg.)}} 
    & \multicolumn{2}{c|}{\textbf{Min $C$ (strong reg.)}} \\

    & RBF & Lin. & RBF & Lin. \\
\hline
2 (linear fit.) & 0 & \textbf{100} & 50 & \textbf{100} \\
2 (second slope + number of $\beta \leq 2$) & 50 & \textbf{90} & \textbf{100} & \textbf{90} \\
2 (second slope + proportion of $\beta \leq 2$) & 50 & 75 & \textbf{100} & \textbf{100} \\
3 (quadratic fit.) & 40 & 50 & \textbf{90} & 70 \\
3 (number of $\beta \leq 2$, $\beta \leq 3$ and $\beta > 3$) & 40 & \textbf{90} & 42.5 & 42.5 \\
3 (proportion of $\beta \leq 2$, $\beta \leq 3$ and $\beta > 3$) & 20 & 40 & 70 & \textbf{90} \\
3 (second slope + number of $\beta \leq 2$ and $\beta \leq 3$) & 40 & 70 & \textbf{90} & 70 \\
3 (second slope + proportion of $\beta \leq 2$ and $\beta \leq 3$) & 50 & 75 & \textbf{100} & \textbf{100} \\
4 (second slope + number of  $\beta \leq 2$, $\beta \leq 3$ and $\beta > 3$) & 40 & 70 & 62.5 & 32.5 \\
4 (second slope + proportion of  $\beta \leq 2$, $\beta \leq 3$ and $\beta > 3$) & 50 & 75 & \textbf{100} & \textbf{100} \\
5 (linear fit. + number of $\beta \leq 2$, $\beta \leq 3$ and $\beta > 3$) & 30 & \textbf{90} & 77.5 & 52.5 \\
5 (linear fit. + proportion of $\beta \leq 2$, $\beta \leq 3$ and $\beta > 3$) & 0 & \textbf{100} & 50 & \textbf{100} \\
6 (quadratic fit. + number of $\beta \leq 2$, $\beta \leq 3$ and $\beta > 3$) & 40 & 72.5 & \textbf{90} & 45 \\
6 (quadratic fit. + proportion of $\beta \leq 2$, $\beta \leq 3$ and $\beta > 3$) & 40 & 62.5 & \textbf{90} & 80 \\
5 (piecewise linear fit.) & 50 & \textbf{100} & \textbf{100} & \textbf{100} \\
8 (piecewise linear fit. + number of $\beta \leq 2$, $\beta \leq 3$ and $\beta > 3$) & 40 & 67.5 & 45 & 45 \\
8 (piecewise linear fit. + proportion of $\beta \leq 2$, $\beta \leq 3$ and $\beta > 3$) & 50 & \textbf{100} & \textbf{100} & \textbf{100} \\
8 (piecewise lin. + quadratic fit.) & 40 & \textbf{100} & \textbf{90} & \textbf{100} \\
13 (lin. fit. + piecewise lin. fit. + quad. fit. + number of $\beta$) & 40 & 80 & \textbf{90} & 70 \\
13 (lin. fit. + piecewise lin. fit. + quad. fit. + proportion of $\beta$) & 30 & \textbf{90} & 80 & \textbf{90} \\
50 (resampled $\pmb{\beta}$ profile) & 40 & \textbf{100} & \textbf{90} & \textbf{100} \\
\hline
\end{tabular}
}
\caption{\small Balanced accuracies of \textbf{SVM models} with an RBF or a linear kernel trained on different sets of features. First box shows results for elevations with \textbf{relative} resampling, second box with \textbf{absolute} resampling. 
Third box shows contour parameters, from \cite{beghin2025identification}. Balanced accuracies $\ge 90\%$ in bold.}
\label{tab:SVM}
\end{table*}

\noindent Elevation maps deliver mixed results in these experiments. Most balanced accuracies are quite low, although some combinations of resolution, kernel and regularisation strategy manage to achieve balanced accuracies reaching 90\% (in bold at the top of Table \ref{tab:SVM}). Counterintuitively, the best results for absolute resamplings are only obtained for the two smallest resolutions, and only for normalised elevations. For relative resamplings, smaller resolutions perform again well with normalised data, but some unnormalised larger grids also reach 90\% accuracy. Kernel choice matters, but the best performing kernel among RBF and Linear seems to very often depend on the other settings, with no clear logic.\\

\noindent The bottom of Table \ref{tab:SVM} displays the results from the parameter profile-based approach in \cite{beghin2025identification}. Note that while some of them rely only on the parameter profile (e.g. in the second line), most of them include secondary features derived from those profiles, such as slopes or values of parameters from a piecewise linear or quadratic fit (see \cite{beghin2025identification} for details). Altogether, the use of engineered features seems to more easily and reliably achieve high balanced accuracies, often reaching 100\%, especially when using strong regularisation. In comparison, it appears that an approach based exclusively on raw elevation maps is not as effective. \\

\noindent We also observe that despite a robust and non-arbitrary choice of the regularisation hyperparameter $C$ in Algorithm \ref{Algo:23-1-1}, both very good and very bad classification results may appear next to each other in Table \ref{tab:SVM}. Recall that a balanced accuracy of 50\% corresponds to random guessing. Values below 50\% indicate that the model performs worse than chance, meaning that its predictions are systematically misleading and could be improved simply by reversing them. We do not fully explain these anomalies, which could be side-effects of the very small size of our corpus. \\

\noindent Reducing the dimensionality of the elevation maps using PCA does not appear to significantly improve the classification performance. When training SVM models on PCA-reduced resampled grids, we observe nearly identical results across all grid resolutions, see Table \ref{tab:svm_rbf_pca} for an RBF kernel and Table \ref{tab:svm_linear_pca} for a linear kernel. Interestingly, a peak balanced accuracy of 90\% is again obtained with fewer features, namely when retaining only two principal components with the RBF kernel and strong regularisation. Using a larger number of retained components does not further improve the results, and even leads to a small decrease in performance. \\

\noindent Overall, the values reported in Tables \ref{tab:svm_rbf_pca} and \ref{tab:svm_linear_pca} for PCA-reduced elevations are not significantly better than those obtained with the raw elevation maps in Table \ref{tab:SVM}, although the best balanced accuracies attained are comparable. This suggests that, in our setting, PCA-based dimensionality reduction does not provide a clear advantage over directly using the resampled grids. Nevertheless, we observe a general trend: as the number of features increases - whether using raw elevation maps or PCA projections - the model tends to become less accurate. This behaviour likely reflects overfitting effects induced by high-dimensional representations in a very small dataset. \\

%
%

\begin{table}
\centering
\resizebox{\textwidth}{!}{
\begin{tabular}{|l|c|c|}
\hline
& \multicolumn{2}{c|}{\textbf{Balanced accuracy} (in $\%$)} \\

\textbf{Number and type of features} 
    & \textbf{Max $C$ (weak reg.)} & \textbf{Min $C$ (strong reg.)} \\
\hline
2 (PCA on all relative grid resampling) & 40 & \textbf{90} \\
3 (PCA on all relative grid resampling) & 20 & 70  \\
5 (PCA on all relative grid resampling) & 20 & 70 \\
8 (PCA on all relative grid resampling) & 20 & 70 \\
10 (PCA on all relative grid resampling) & 20 & 70 \\
12 (PCA on all relative grid resampling) & 20 & 70 \\
\hline
\end{tabular}
}
\caption{\small Balanced accuracies of SVM models with an RBF kernel trained on different PCA projections of the relative resampled grids. Balanced accuracies $\ge 90\%$  in bold.}
\label{tab:svm_rbf_pca}
\end{table}

%
%

\begin{table}
\centering
\resizebox{\textwidth}{!}{
\begin{tabular}{|l|c|c|}
\hline
& \multicolumn{2}{c|}{\textbf{Balanced accuracy} (in $\%$)} \\

\textbf{Number and type of features} 
    & \textbf{Max $C$ (weak reg.)} & \textbf{Min $C$ (strong reg.)} \\
\hline
2 (PCA on $5 \times 10$ relative grid resampling) & 72.5 & 72.5 \\
3 (PCA on $5 \times 10$ relative grid resampling) & 72.5 & 72.5  \\
10 (PCA on $5 \times 10$ relative grid resampling) & 72.5 & 72.5 \\

\hline
\end{tabular}
}
\caption{\small Balanced accuracies of SVM models with a linear kernel trained on different PCA projections of the relative resampled grids. Balanced accuracies $\ge 90\%$ in bold.}
\label{tab:svm_linear_pca}
\end{table}

\subsection{Decision Trees classifiers}

\noindent We now turn to the results obtained with Decision Trees. As for the SVM models, both the \textit{relative} and \textit{absolute} resampling approaches were evaluated, and Algorithm \ref{Algo:23-1-1} was used to determine the optimal tree depth in order to avoid favouring arbitrary fixed depths. Since Decision Trees are not distance-based models, no normalisation of the elevation maps is required. We also compared two splitting criteria: Gini impurity and Shannon entropy. \\

\noindent Overall, Decision Trees yield weaker performance than SVMs. The classification performance remains modest for both the raw grids (top of Table~\ref{tab:Decision Trees}) and their PCA projections (Table~\ref{tab:decision_tree_pca}). Slightly stronger predictions are again observed when using features derived from the $\beta$ parameter profiles (bottom of Table~\ref{tab:Decision Trees}). This confirms that problem-specific, low-dimensional geometric descriptors remain more informative than generic grid-based representations for the detection of reduced instruments in our dataset.

%
%

\begin{table*}
\centering
\resizebox{\textwidth}{!}{
\begin{tabular}{|l|cc|cc|}
\hline
& \multicolumn{4}{c|}{\textbf{Balanced accuracy} (in $\%$)} \\
\textbf{Number and type of features} & \multicolumn{2}{c|}{\textbf{min Depth}} & \multicolumn{2}{c|}{\textbf{max Depth}} \\
            & Gini & Entropy & Gini & Entropy \\
\hline
50 ($5 \times 10$ relative grid resampling)       & 35 & 40 & 42.5 & 40 \\
250 ($10 \times 25$ relative grid resampling)     & 40 & 37.5 & 45 & 37.5 \\
1000 ($20 \times 50$ relative grid resampling)    & 45 & 37.5 & 45 & 35 \\
1625 ($25 \times 65$ relative grid resampling)    & 40 & 32.5 & 40 & 40 \\
6250 ($50 \times 125$ relative grid resampling)   & 37.5 & 35 & 35 & 32.5 \\
14250 ($75 \times 190$ relative grid resampling)  & 30 & 42.5 & 30 & 40 \\
25000 ($100 \times 250$ relative grid resampling) & 52.5 & 55 & 55 & 55 \\
\hline
50 ($5 \times 10$ absolute grid resampling)       & 52.5 & 65 & 65 & 47.5 \\
250 ($10 \times 25$ absolute grid resampling)     & 82.5 & 65 & 75 & 62.5 \\
1000 ($20 \times 50$ absolute grid resampling)    & 70 & 77.5 & 80 & 77.5 \\
1625 ($25 \times 65$ absolute grid resampling)    & 82.5 & 82.5 & 80 & 77.5 \\
6250 ($50 \times 125$ absolute grid resampling)   & 55 & 37.5 & 55 & 42.5 \\
14250 ($75 \times 190$ absolute grid resampling)  & 62.5 & 52.5 & 65 & 52.5 \\
25000 ($100 \times 250$ absolute grid resampling) & 70 & 50 & 72.5 & 50 \\
\hline
\end{tabular}
}

\vspace*{.2cm}

\resizebox{\textwidth}{!}{
\begin{tabular}{|l|cc|cc|}
\hline
& \multicolumn{4}{c|}{\textbf{Balanced accuracy} (in $\%$)} \\
\textbf{Number and type of features} & \multicolumn{2}{c|}{\textbf{min Depth}} & \multicolumn{2}{c|}{\textbf{max Depth}} \\
            & Gini & Entropy & Gini & Entropy \\
\hline
2 (linear fit.) & 40 & 35 & 30 & 40 \\
2 (second slope + number of $\beta \leq 2$) & 72.5 & 82.5 & 52.5 & 72.5 \\
2 (second slope + proportion of $\beta \leq 2$) & 82.5 & \textbf{92.5} & 72.5 & 82.5 \\
3 (quadratic fit.) & 60 & 50 & 60 & 52.5 \\
3 (number of $\beta \leq 2$, $\beta \leq 3$ and $\beta > 3$) & 82.5 & 82.5 & 82.5 & 82.5 \\
3 (proportion of $\beta \leq 2$, $\beta \leq 3$ and $\beta > 3$) & 80 & 80 & 80 & 80 \\
3 (second slope + number of $\beta \leq 2$ and $\beta \leq 3$) & 82.5 & 82.5 & 62.5 & 72.5 \\
3 (second slope + proportion of $\beta \leq 2$ and $\beta \leq 3$) & 82.5 & \textbf{92.5} & 72.5 & 72.5 \\
4 (second slope + number of  $\beta \leq 2$, $\beta \leq 3$ and $\beta > 3$) & 82.5 & 82.5 & 72.5 & 72.5 \\
4 (second slope + proportion of  $\beta \leq 2$, $\beta \leq 3$ and $\beta > 3$) & 82.5 & 82.5 & 62.5 & 72.5 \\
5 (linear fit. + number of $\beta \leq 2$, $\beta \leq 3$ and $\beta > 3$) & 82.5 & 82.5 & 82.5 & 82.5 \\
5 (linear fit. + proportion of $\beta \leq 2$, $\beta \leq 3$ and $\beta > 3$) & 82.5 & 82.5 & \textbf{92.5} & 82.5 \\
6 (quadratic fit. + number of $\beta \leq 2$, $\beta \leq 3$ and $\beta > 3$) & 82.5 & 82.5 & 82.5 & 82.5 \\
6 (quadratic fit. + proportion of $\beta \leq 2$, $\beta \leq 3$ and $\beta > 3$) & \textbf{92.5} & \textbf{92.5} & \textbf{92.5} & \textbf{92.5} \\
5 (piecewise linear fit.) & 62.5 & 62.5 & 62.5 & 62.5 \\
8 (piecewise linear fit. + number of $\beta \leq 2$, $\beta \leq 3$ and $\beta > 3$) & 75 & 65 & 75 & 65 \\
8 (piecewise linear fit. + proportion of $\beta \leq 2$, $\beta \leq 3$ and $\beta > 3$) & 82.5 & 72.5 & 82.5 & 72.5 \\
8 (piecewise lin. + quadratic fit.) & 62.5 & 62.5 & 62.5 & 62.5 \\
13 (lin. fit. + piecewise lin. fit. + quad. fit. + number of $\beta$) & 75 & 75 & 75 & 75 \\
13 (lin. fit. + piecewise lin. fit. + quad. fit. + proportion of $\beta$) & 82.5 & 82.5 & 82.5 & 82.5 \\
50 (resampled $\pmb{\beta}$ profile) & 70 & 52.5 & 72.5 & 60 \\
\hline
\end{tabular}
}
\caption{\small Balanced accuracies of \textbf{Decision Trees models} with a \texttt{gini} or \texttt{entropy} criterion trained on different sets of features. First box shows results for elevations with \textbf{relative} resampling, second box with \textbf{absolute} resampling. Third box shows contour parameters, from \cite{beghin2025identification}. Balanced accuracies $\ge 90\%$ in bold}
\label{tab:Decision Trees}
\end{table*}

%
%

\begin{table}
\centering
\resizebox{\textwidth}{!}{
\begin{tabular}{|l|cc|cc|}
\hline
& \multicolumn{4}{c|}{\textbf{Balanced accuracy} (in $\%$)} \\

\textbf{Number and type of features} 
    & \multicolumn{2}{c|}{\textbf{min Depth}} 
    & \multicolumn{2}{c|}{\textbf{max Depth}} \\

    & Gini & Entropy & Gini & Entropy \\
\hline
2 (PCA on 5 x 10 relative grid resampling) & 67.5 & 67.5 & 67.5 & 67.5 \\
2 (PCA on 10 x 25 relative grid resampling) & 55 & 65 & 52.5 & 62.5 \\
2 (PCA on 20 x 50 relative grid resampling) & 55 & 65 & 55 & 65 \\
2 (PCA on 25 x 65 relative grid resampling) & 55 & 65 & 55 & 65 \\
2 (PCA on 50 x 125 relative grid resampling) & 62.5 & 62.5 & 62.5 & 62.5 \\
2 (PCA on 75 x 190 relative grid resampling) & 52.5 & 62.5 & 62.5 & 52.5 \\
2 (PCA on 100 x 250 relative grid resampling) & 52.5 & 52.5 & 52.5 & 52.5 \\
\hline
3 (PCA on 5 x 10 relative grid resampling) & 57.5 & 67.5 & 67.5 & 57.5 \\
3 (PCA on 10 x 25 relative grid resampling) & 45 & 50 & 45 & 50 \\
3 (PCA on 20 x 50 relative grid resampling) & 45 & 52.5 & 45 & 52.5 \\
3 (PCA on 25 x 65 relative grid resampling) & 45 & 52.5 & 45 & 52.5 \\
3 (PCA on 50 x 125 relative grid resampling) & 62.5 & 62.5 & 62.5 & 62.5 \\
3 (PCA on 75 x 190 relative grid resampling) & 62.5 & 62.5 & 52.5 & 62.5 \\
3 (PCA on 100 x 250 relative grid resampling) & 52.5 & 32.5 & 32.5 & 35 \\
\hline
5 (PCA on 5 x 10 relative grid resampling) & 57.5 & 70 & 47.5 & 60 \\
5 (PCA on 10 x 25 relative grid resampling) & 32.5 & 42.5 & 32.5 & 42.5 \\
5 (PCA on 20 x 50 relative grid resampling) & 32.5 & 42.5 & 32.5 & 42.5 \\
5 (PCA on 25 x 65 relative grid resampling) & 42.5 & 52.5 & 42.5 & 42.5 \\
5 (PCA on 50 x 125 relative grid resampling) & 62.5 & 62.5 & 62.5 & 62.5 \\
5 (PCA on 75 x 190 relative grid resampling) & 62.5 & 52.5 & 52.5 & 52.5 \\
5 (PCA on 100 x 250 relative grid resampling) & 55 & 37.5 & 35 & 40 \\
\hline
8 (PCA on 5 x 10 relative grid resampling) & 47.5 & 37.5 & 47.5 & 47.5 \\
8 (PCA on 10 x 25 relative grid resampling) & 50 & 65 & 52.5 & 62.5 \\
8 (PCA on 20 x 50 relative grid resampling) & 72.5 & 72.5 & 72.5 & 72.5 \\
8 (PCA on 25 x 65 relative grid resampling) & 35 & 45 & 45 & 35 \\
8 (PCA on 50 x 125 relative grid resampling) & 55 & 62.5 & 55 & 60 \\
8 (PCA on 75 x 190 relative grid resampling) & 55 & 62.5 & 55 & 60 \\
8 (PCA on 100 x 250 relative grid resampling) & 55 & 60 & 55 & 47.5 \\
\hline
10 (PCA on 5 x 10 relative grid resampling) & 57.5 & 37.5 & 47.5 & 47.5 \\
10 (PCA on 10 x 25 relative grid resampling) & 50 & 52.5 & 52.5 & 62.5 \\
10 (PCA on 20 x 50 relative grid resampling) & 62.5 & 52.5 & 52.5 & 52.5 \\
10 (PCA on 25 x 65 relative grid resampling) & 45 & 42.5 & 45 & 55 \\
10 (PCA on 50 x 125 relative grid resampling) & 55 & 62.5 & 55 & 62.5 \\
10 (PCA on 75 x 190 relative grid resampling) & 45 & 52.5 & 45 & 62.5 \\
10 (PCA on 100 x 250 relative grid resampling) & 55 & 47.5 & 55 & 47.5 \\
\hline
12 (PCA on 5 x 10 relative grid resampling) & 47.5 & 47.5 & 47.5 & 37.5 \\
12 (PCA on 10 x 25 relative grid resampling) & 50 & 62.5 & 52.5 & 62.5 \\
12 (PCA on 20 x 50 relative grid resampling) & 50 & 60 & 50 & 50 \\
12 (PCA on 25 x 65 relative grid resampling) & 45 & 42.5 & 45 & 45 \\
12 (PCA on 50 x 125 relative grid resampling) & 65 & 62.5 & 55 & 62.5 \\
12 (PCA on 75 x 190 relative grid resampling) & 65 & 52.5 & 55 & 62.5 \\
12 (PCA on 100 x 250 relative grid resampling) & 52.5 & 50 & 52.5 & 60 \\
\hline
\end{tabular}
}
\caption{\small Balanced accuracies of \textbf{Decision Trees models} with a \texttt{gini} or \texttt{entropy} criterion trained on different PCA projection of the relative resampled grids.}
\label{tab:decision_tree_pca}
\end{table}

\section{Conclusions and Perspectives}

In this work, we have studied the possibility of detecting reduced violins using classifiers applied to their raw elevation maps. Despite their higher dimension (up to 25 000 features) and hence their potential for containing more useful information, it turns out that their use with robust SVM regularisation or with Decision Tree depth validation does not consistently match the performance of the feature-engineering approach based on contour line parameters, although they sometimes reach comparable accuracy. Nevertheless, although elevation maps cannot fully replace contour lines parameters, they can still complement them by providing a more complete description of the the arching and shape of the sound board.

\smallskip  \noindent Classification of reduced violins is a complex problem involving a small corpus of data. We restricted ourselves to sound boards of violins and violas. Future research will incorporate violin backs and integrate cello datasets. We have observed that classical methods used to classify 3D point clouds are ineffective, and that problem-specific feature engineering can improve results of a SVM classifier compared to raw elevation maps. In future work we plan to further explore a combination of those features, as well as revisit the use of multilayer perceptrons or convolutional neural networks.

%
%

\bibliographystyle{unsrt}
\bibliography{refs}
\end{document}